# An Integrated Autoencoder-Based Filter for Sparse Big Data

Wei Peng[1], Baogui Xin[1*]

[1] College of Economics and Management, Shandong University of Science and Technology, Qingdao 266590, China.

**\* Correspondence:**
Corresponding Author
xin@tju.edu.cn

**Abstract**

We propose a novel filter for sparse big data, called an integrated autoencoder (IAE), which utilizes auxiliary information to mitigate data sparsity. The proposed model achieves an appropriate balance between prediction accuracy, convergence speed, and complexity. We implement experiments on a GPS trajectory dataset, and the results demonstrate that the IAE is more accurate and robust than some state-of-the-art methods.

## 1 Introduction

### 1.1 Motivation

Prediction is essential in many scientific fields, such as economics [1-7], engineering [8-12], and medicine [13-16]. A valid prediction can facilitate sound decisions. But high sparsity of original data can substantially decrease prediction performance. Existing prediction algorithms [17-21] are not good at processing such data. We may leverage auxiliary information to mitigate the sparsity of data, but these algorithms have difficulty representing complicated features when auxiliary data are unstructured and heterogeneous. The deep learning-based scheme can extract complex nonlinear representations from these data, and it usually shows better prediction performance.

Deep learning has led to remarkable achievements in fields such as computer vision [22-25], speech recognition [26-30], and natural language processing [31, 32], which have attracted much attention in theory and practice. Many deep learning frameworks have been developed, such as the convolutional neural network (CNN) [24, 33-35], recurrent neural network (RNN) [36-39], long short-term memory (LSTM) [1, 32, 40, 41], deep belief network (DBN) [6, 8, 42, 43], AE [9, 44-47] and so on. In some circumstances, deep learning frameworks deal inefficiently with unstructured and heterogeneous data, because data preprocessing increases the complexity of the structure.

The AE has low complexity and no data preprocessing requirement. AEs are widely applied in prediction [2, 9, 11, 48], classification [49-52], and latent representation learning [44, 45, 53, 54]. Various requirements give rise to many variations of AEs. Vincent et al. [44, 45] proposed a denoising AE (DAE), stack AEs (SAE), and stack denoising AEs (SDAE), which focus on extracting more abstract representations for a standalone supervised deep learning architecture. Görgel and Simsek [12] developed deep stacked denoising sparse autoencoders (DSDSA) for face recognition and the sparsity property of DSDSA substantially decreased the possibility of overfitting. Lv et al. [55] applied an SAE to tourism demand prediction using search query data, which has a better performance than some other state-of-the-art deep learning-based models. The SAE and SDAE are stacked by the AE and DAE and obtain their optimal values through several optimization processes, which lead to low efficiency. In brief, the above variations of the AE obtain higher accuracy by increasing complexity and decreasing efficiency.

## 1.2 Contributions

Autoencoder (AE), stacked AE (SAE), denoising AE (DAE) and stacked DAE (SDAE) are efficient prediction approaches with low complexity that has been widely applied in some scientific fields. However, to obtain a well-trained SAE or SDAE stacked by $k$ AEs or DAEs, we must optimize $k+1$ objective functions, which require a significant amount of computing resources. By integrating autoencoders, we develop a novel filter only by optimizing a single objective function. In addition, the integrated autoencoder (IAE) utilize auxiliary information to mitigate data sparsity to achieve an appropriate balance among prediction accuracy, convergence speed, and complexity. Experiments demonstrate that the proposed IAE performs superior than above-mentioned model in accuracy and efficiency.

## 1.3 Framework

The remainder of this paper is organized as follows. Section 2 reviews some related prediction methods. Section 3 describes the IAE in detail. Section 4 discusses experiments and results. Section 5 summarizes conclusions.

## 2 Related Work

### 2.1 Non-AE prediction methods

Machine learning-based prediction methods can effectively deal with nonlinear prediction problems. They consist mainly of singular value decomposition (SVD) [56, 57], matrix factorization (MF) [18, 58], support vector machine (SVM) [59, 60], and Bayesian models [61, 62]. Machine learning-based prediction can enhance prediction performance to some extent. However, it is hard to obtain a substantial breakthrough.

Deep learning-based methods are typically composed of numerous neurons and parameters, which can extract more complicated and nonlinear features from the input data. For example, Krizhevsky et al. [24] proposed a deep CNN with 60 million parameters and 650,000 neurons to classify 1000 image categories. Since the dataset they employed contained 1.2 million images, the deep CNN architecture could prevent overfitting to some extent and performed better than previous deep learning-based methods. However, it was too complicated to train quickly. Ting et al. [33] proposed a CNN-based architecture (CNNI-BCC) for breast cancer classification. Compared with previous work, CNNI-BCC is indeed more accurate. However, it contains 30 hidden layers and each layer contains lots of parameters, which leads to low efficiency. Chan et al. [63] proposed a simpler unsupervised convolutional neural network (PCANet) with fewer parameters, which could accelerate training, but whose performance was significantly degraded when the images were not well prepared. Song et al. [64] developed a content-based automatic tagging algorithm with five-layer RNNs which performed a higher training efficiency than above-mentioned algorithms with numerous layers and parameters. But the model need a preprocessing phase, which also takes too much time. LSTM, as an extension of RNN, is a more sophisticated but effective structure with a gating mechanism. Baek and Kim [1] proposed a data-augmentation approach (ModAugNet) consisting of an overfitting prevention LSTM and a prediction LSTM, which can avoid overfitting by overcoming the deficiency of limited data availability. Mohamed et al. [65] replaced Gaussian mixture models by DBNs for acoustic modelling and achieved a better performance. But the architecture contained lots of hidden neurons and parameters in every layer, which could impact its convergence speed.

## 2.2 AE and its Variations

Though the above methods can significantly improve prediction performance, most are complicated, which may cause low efficiency. The AE is a simple but effective deep learning approach that can learn complex representations with high efficiency.

### 2.2.1 *AE*

An AE is an unsupervised representation learning framework with three parts: the input layer, hidden layer, and output layer, which are defined as

$$H = f_\theta(x) = f(W \times x + b) \quad (1)$$

$$O = g_{\theta'}(H) = g(W' \times H + b') \quad (2)$$

where $x$ is the original input; $H$ is the hidden representation; $O$ is the output; $f(\cdot)$ and $g(\cdot)$ are encoder and decoder activation functions, respectively; $W$ and $W'$ are weights; $b$ and $b'$ are biases; $\theta = \{W, b\}$; and $\theta' = \{W', b'\}$. Figure 1 describes the architecture of an AE.

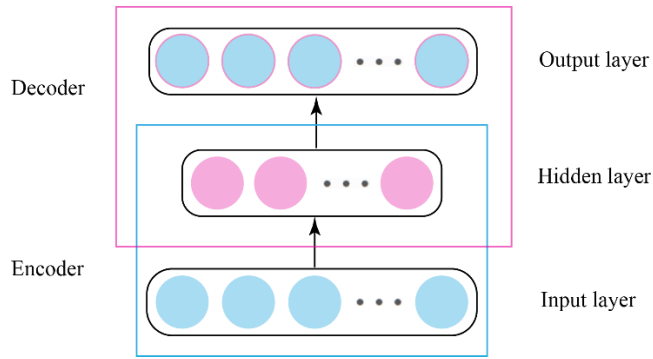

Figure 1: Architecture of an AE.

To obtain optimal parameters by minimizing the error between input and output layers, we write the following loss function of the AE:

$$e_{AE} = min \sum \left\| x - g\left(W' \times f(W \times x + b) + b'\right) \right\|_\mathcal{O}^2 \quad (3)$$

where $\left\| \cdot \right\|_\mathcal{O}^2$ denotes that we consider only the observed data.

The AE can learn latent representations by simply copying input, but the learned representations are usually too specific to generalize well. In some scenarios, one must enhance the generalization ability by stacking several AEs, which can learn more representative features than a single AE. Also, for different demands, one must propose some variations of the AE.

Thus some researchers have combined the AE with other deep learning models to both improve efficiency and accuracy. AE-based methods perform better with sparse data, such as rating data in recommender systems. Sedhain et al. [66] proposed an AE-based collaborative model (AutoRec), which has advantages over existing neural network approaches in representing and computing. Lv et al. [9] applied an SAE to traffic prediction, which is indispensable to better city planning.

### 2.2.2 Stacked AE (SAE)

An SAE stacks several typical AEs to attain stronger generalization ability. Its training process includes pre-training and fine-tuning, as shown in Figure 2.

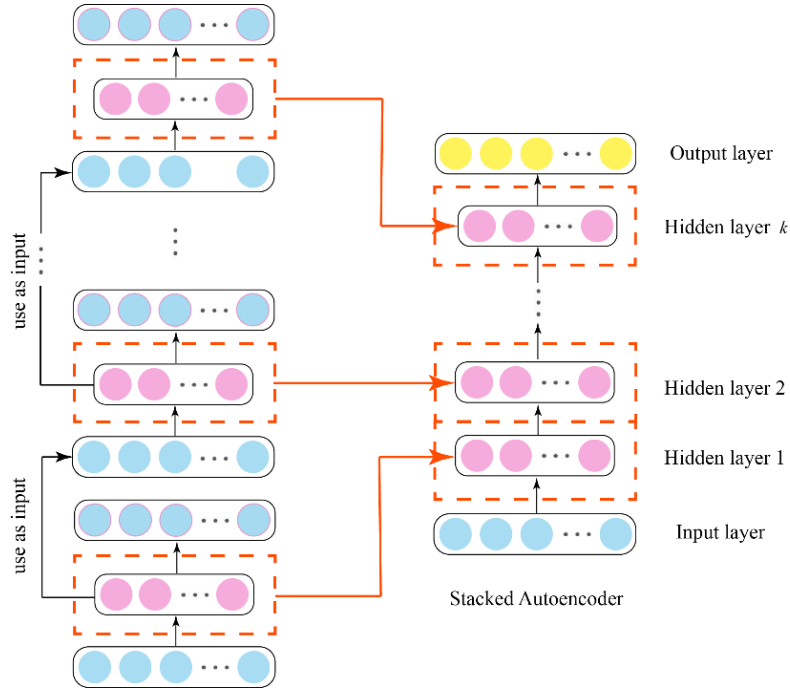

Figure 2: Training process of SAE.

According to Figure 2, assuming there is an SAE stacked by $k$ AEs, the output of each AE in the pre-training stage is discarded, and each hidden layer is used as the input of the next AE until the $k^{th}$ AE is well trained. After pre-training each AE, the original input is transformed to the top AE, layer by layer. The hidden layer of each AE can be calculated by

$$H_i = \begin{cases} g^{(1)}\left(W_1' \times f^{(1)}\left(W_1 \times x + b_1\right) + b_1'\right), & i = 1 \\ g^{(i)}\left(W_i' \times f^{(i)}\left(W_i \times H_{i-1} + b_i\right) + b_i'\right), & i = 2, \cdots, k \end{cases} \quad (4)$$

where $H_i$ represents the hidden layer of the $i^{th}$ AE; $f^{(i)}$ and $g^{(i)}$ respectively represent the encoder and decoder activation functions of the $i^{th}$ AE; $W_i$ and $W_i'$ respectively represent the encoder and decoder weight of the $i^{th}$ AE; and $b_i$ and $b_i'$ respectively represent the encoder and decoder bias of the $i^{th}$ AE.

The top hidden layer of the SAE can be taken as input to a standalone supervised learning approach. For example, to finish classification tasks, one often adds a softmax classification layer connected to the top hidden layer. In the fine-tuning stage, to obtain the fine-tuned parameters by minimizing the reconstruction error, the loss function is defined by

$$e_{SAE} = min\left\{\|x - \hat{x}\|_{\mathcal{O}}^2, \left\{min\left(e_{AE}^{(k)}\right), \left\{min\left(e_{AE}^{(k-1)}\right), \left\{\cdots, \left|min\left(e_{AE}^{(1)}\right)\right.\right\}\right\}\right\}\right\} \quad (5)$$

where $e_{AE}^{(k)}$ denotes the reconstruction error of the $k^{th}$ AE.

Though the SAE has better generalization ability, its performance can be affected when the original input is mixed with noise. In some circumstances, an AE needs a denoised input.

### 2.2.3 Denoising AE (DAE)

A DAE reconstructs a repaired (or "clean") input from a partially destroyed (or "corrupted") one. It is achieved by first partially destroying the raw input $x$ to obtain a corrupted version $\tilde{x}$ via a stochastic mapping $\tilde{x} \sim q_D(\tilde{x}|x)$. And then $\tilde{x}$ is mapped by an AE, which includes the encoding process $h = f_\theta(\tilde{x}) = f(W\tilde{x} + b)$ and the decoding process $\hat{x} = g_{\theta'}(y) = g(W'h + b')$. Figure 3 illustrates the mechanism of a DAE.

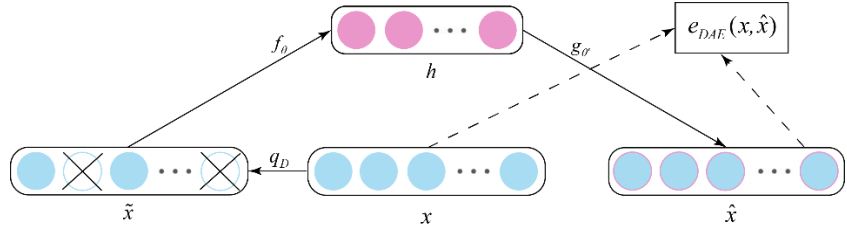

Figure 3: Mechanism of a DAE.

Similarly, parameters are optimized by the following loss function, which minimizes the reconstruction error between the raw input $x$ and output $z$:

$$e_{DAE} = min\sum \|x - g(W' \times f(W \times \tilde{x} + b) + b')\|_{\mathcal{O}}^2 \quad (6)$$

Indeed, it is natural to take into account stacking DAEs to obtain a more powerful AE-based model by exploiting their respective advantages.

### 2.2.4 Stacking DAE (SDAE)

Stacking DAEs or typical AEs to build a deep network is the same as stacking RBMs in deep belief networks. The training process of an SDAE also includes pre-training and fine-tuning. The structure of an SDAE is shown in Figure 4.

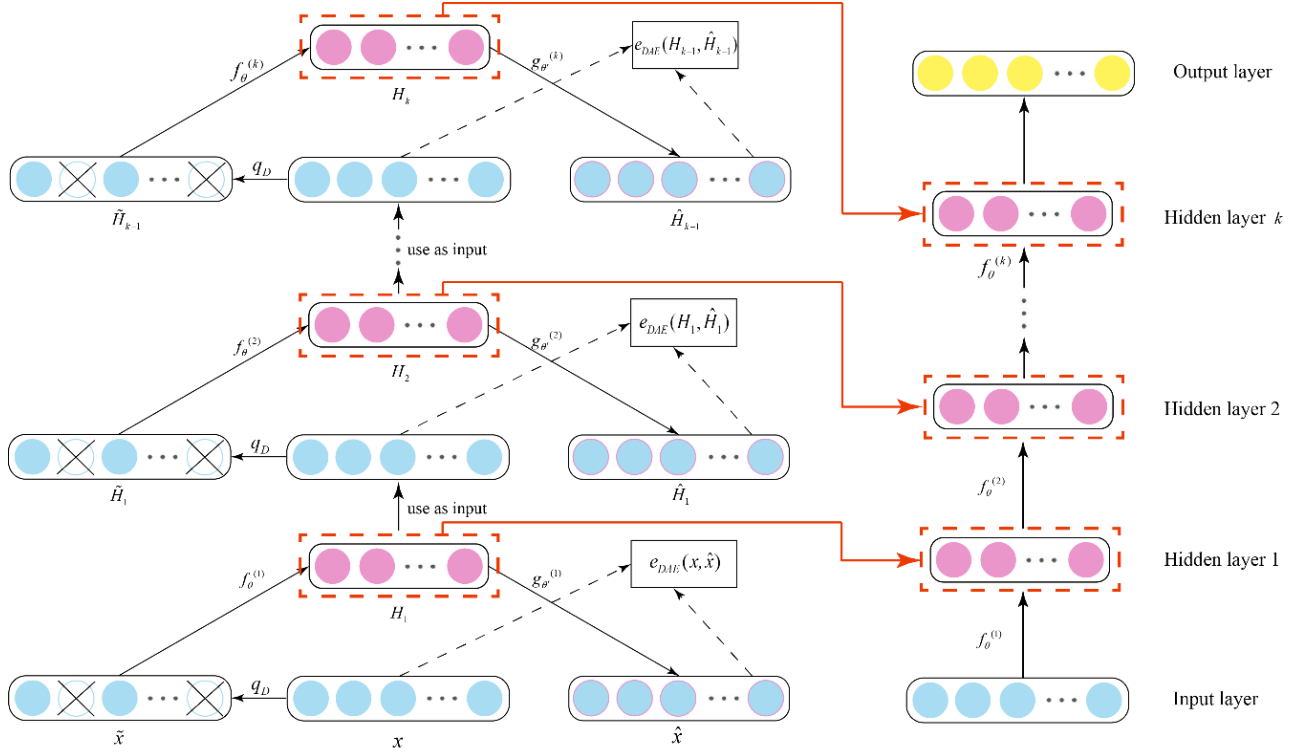

Figure 4: Training process of SDAE.

In the pre-training stage, the output of each DAE is dropped, and each hidden layer is used as the input of the next DAE until the $k^{th}$ DAE is well trained. Similarly, to finish classification tasks, one often adds a softmax classification layer as an output layer connected to the top hidden layer. In the fine-tuning stage, to obtain the fine-tuned parameters by minimizing the reconstruction error, the loss function is defined by

$$e_{SDAE} = min\left\{\|x-\hat{x}\|_{\mathcal{O}}^2, \left\{min\left(e_{DAE}^{(k)}\right), \left\{min\left(e_{DAE}^{(k-1)}\right), \left\{\cdots, min\left(e_{DAE}^{(1)}\right)\right\}\right\}\right\}\right\} \qquad (7)$$

where $e_{DAE}^{(k)}$ denotes the reconstruction error of the $k^{th}$ DAE.

There are some previous works based on above-mentioned SAE and SDAE. Tong et al. [11] developed a novel architecture which makes use of SDAEs and ensemble learning (SDAEsTSE) for software defect prediction. And the SDAEsTSE achieves a good balance between accuracy and efficiency. Bai et al. [67] proposed a DL-SSAE model which takes into account seasonality to predict PM 2.5 concentration. And the experimental results demonstrate that the DL-SSAE outperforms some state-of-the-art models whether or not to consider seasonality. To sum up, the AE is an efficient prediction approach with low complexity that has been widely applied in some scientific fields.

### 2.2.5 Summary

There is no doubt that SAE and SDAE can improve prediction accuracy. However, to obtain a well-trained SAE or SDAE stacked by $k$ AEs or DAEs, we must optimize $k+1$ objective functions, which will consume a lot of computing resources. Therefore, it is urgent to develop a new variation of AE to overcome the shortcoming caused by multiply objective functions. In the next Section, we will develop a novel filter only by optimizing a single objective function.

## 3      Integrated autoencoder (IAE)

In this section, we introduce a novel filter with a single objective function for sparse big data, called IAE, which utilizes auxiliary information to mitigate data sparsity. The IAE model joins two AEs to enhance both accuracy and efficiency. The structure of the IAE is depicted in Figure 5.

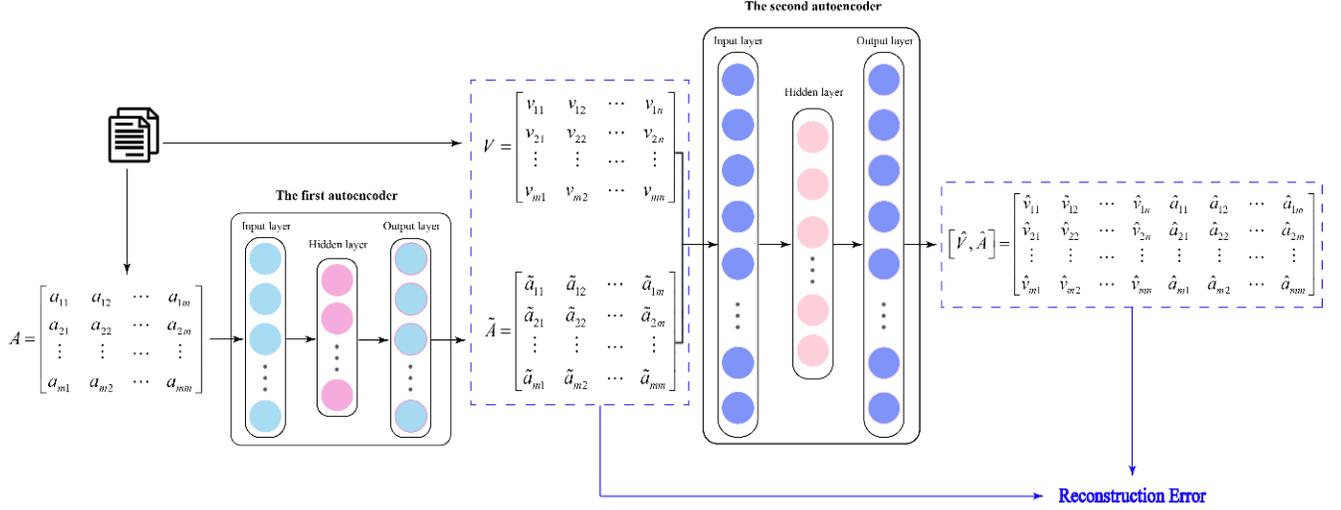

Figure 5: Structure of the proposed IAE model.

Assume that there is a matrix $V \in \mathbb{R}^{m \times n}$ that must be predicted, and an auxiliary information matrix $A \in \mathbb{R}^{m \times m}$ that can be employed to enhance the prediction accuracy of $V$.

At the first stage, we feed $A$ as the input of the first AE, which will perform representation learning from $A$ and output a reduction version $\tilde{A}$.

$$\tilde{A} = g^{(1)}\left(W_1' \times f^{(1)}\left(W_1 \times A + b_1\right) + b_1'\right) \tag{8}$$

At the second stage, we concatenate $V$ and $\tilde{A}$ horizontally to generate a new matrix $[V, \tilde{A}]$ which is fed as the input of the second AE. Similarly, the second AE performs representation learning from $[V, \tilde{A}]$ and derives a predicted matrix $[\hat{V}, \hat{A}]$. We take the first $n$ columns of $[\hat{V}, \hat{A}]$ as the predicted matrix $\hat{V}$.

Also, we leverage the back-propagation algorithm to minimize the reconstruction error between $V$ and $\hat{V}$ at the second stage. To prevent overfitting, we add the regularization term to the following loss function:

$$e_{JAE} = min \sum \left\|[V, \tilde{A}] - g^{(2)}\left(W_2' \times f^{(2)}\left(W_2 \times [V, \tilde{A}] + b_2\right) + b_2'\right)\right\|_\mathcal{O}^2 + \frac{\lambda}{2}\left(\|W_2\|_F^2 + \|W_2'\|_F^2\right) \tag{9}$$

where $\lambda$ is a hyper parameter.

In the IAE, the second AE implements the error minimization between $[V, \tilde{A}]$ and $[\hat{V}, \hat{A}]$, while the first AE does not perform the error minimization between $A$ and $\tilde{A}$. What is more, there are only two hidden layers in an IAE, and it needs no pre-processing. So, the IAE must consume less running time than the SAE, DAE, or SDAE.

## 4 Experiments

### 4.1 Experimental setup

#### 4.1.1 *Datasets*

We conducted experiments on the GPS trajectory dataset, which contains real-time traffic conditions and POI features of Beijing. Our purpose is to take POI (point of interest) features as auxiliary information to predict the traffic conditions $\bar{v}_{r,t}$ and $sdv_{r,t}$, which respectively denote the average travel speed and standard deviation of the travel speed of all of the vehicles traversing road segment $r$ in the time slot $t$.

The GPS trajectory dataset also has been employed [68]; this is a context-aware matrix factorization method to predict gas consumption and pollution emissions via the prediction of $\bar{v}$ and $sdv$. Moreover, it can be widely used for urban computing scenarios [69].

To verify the performance of the IAE, we randomly divided the experimental dataset into two parts, a training dataset $D_{train}$ and testing dataset $D_{test}$, accounting for $\varphi\%$ and $1-\varphi\%$, respectively.

#### 4.1.2 *Evaluation Index*

Root mean square error (RMSE) is a widely used evaluation index for recommender systems or continuous type prediction. By calculating the RMSE between observed values and predicted values on $D_{test}$, we can test the accuracy of the IAE. A smaller RMSE value represents better performance. It is defined as

$$RMSE = \sqrt{\frac{\sum_{(r,t) \in D_{test}} \left(\bar{v}_{r,t} - \hat{\bar{v}}_{r,t}\right)^2}{|D_{test}|}} \tag{10}$$

where $\hat{\bar{v}}_{r,t}$ represents the predicted value of $\bar{v}_{r,t}$, and $|\cdot|$ represents the number of elements in a set.

The efficiency is measured by the average time $A_t$ consumed in each epoch of the program:

$$A_t = \frac{T_t}{n_{epochs}} \tag{11}$$

where $T_t$ denotes the total time consumed in the program, and $n_{epochs}$ denotes the number of epochs in the program.

#### 4.1.3 *Experimental Implementation and Parameter Settings*

We performed all experiments in Python on a PC with a 2.00 GHz CPU, 16 GB memory, and an NVIDIA GTX 1080Ti-11G GPU.

We adopted the Adam optimizer to minimize the reconstruction error and set some important parameters, as shown in Table 1.

Table 1: Parameter settings.

|  | Parameters | $\bar{v}$ | $sdv$ |
|---|---|---|---|
| Common parameters | Hyper-parameter | $\lambda=0.01$ | $\lambda=0.01$ |
|  | Training ratio | $\varphi=80$ | $\varphi=80$ |
|  | Training epochs | $n=100$ | $n=100$ |
| AE | Number of neurons in hidden layer | 1,000 | 1,000 |
| SAE | Number of neurons in first hidden layer | 1,000 | 1,000 |
|  | Number of neurons in second hidden layer | 3,000 | 3,000 |
| DAE | Number of neurons in hidden layer | 1,000 | 1,000 |
| SDAE | Number of neurons in first hidden layer | 1,000 | 1,000 |
|  | Number of neurons in second hidden layer | 3,000 | 3,000 |
| IAE | Number of neurons in first hidden layer | 1,000 | 1,000 |
|  | Number of neurons in second hidden layer | 3,000 | 3,000 |

## 4.2 Experimental Results

In this section, we set the training epochs of different AE-based algorithms to 100, and obtained the corresponding results, as shown in Table 2.

Table 2: Performance Comparison of Algorithms.

| Algorithm | RMSE and $A_t$ | $\bar{v}$ | $sdv$ |
|---|---|---|---|
| AE | RMSE | 3.968 | 1.374 |
|  | $A_t$ | 4.18 s | 3.92 s |
| SAE | RMSE | 4.294 | 1.511 |
|  | $A_t$ | 33.52 s | 35.75 s |
| DAE | RMSE | 4.054 | 1.444 |
|  | $A_t$ | 4.14 s | 4.77 s |
| SDAE | RMSE | 4.299 | 1.508 |
|  | $A_t$ | 29.95 s | 34.58 s |
| IAE | RMSE | 2.079 | 0.455 |
|  | $A_t$ | 23.83 s | 21.44 s |

Table 2 shows that the accuracy of the IAE exceeds that of other AE-based models. The efficiency of the IAE is higher than that of the SAE and SDAE. It is inevitable that the speed of the AE and DAE is higher than that of the IAE, because the AE and DAE consist of a single hidden layer while the SAE, SDAE, and IAE consist of at least two hidden layers (in this paper, we consider only the SAE, SDAE, and IAE with two hidden layers). Moreover, the SAE, DAE, and SDAE focus on extracting more abstract representations for a standalone supervised deep learning architecture. Therefore, it is justifiable that the RMSEs of the SAE, DAE, and SDAE are all higher than that of the AE.

Figures 6 and 7 show the trend that the accuracy of different algorithms varies with the increase of training epochs. From Figure 6, we note that the RMSEs of the IAE continue decreasing between about epoch 28 and epoch 76, while those of the AE, DAE, SAE, and SDAE decrease at the first several epochs but later fluctuate in a broad range. From Figure 7, we find that the RMSEs of the IAE continue to decrease with increased training epochs, while those of the AE, DAE, SAE, and SDAE continue fluctuating and do not show a significant downward trend. It is clear that the IAE is more accurate and robust than the AE, DAE, SAE, and SDAE.

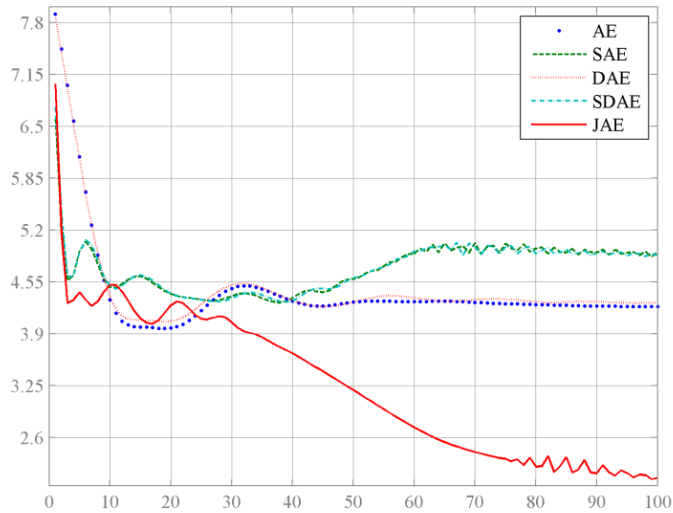

Figure 6: RMSE of $\bar{v}$.

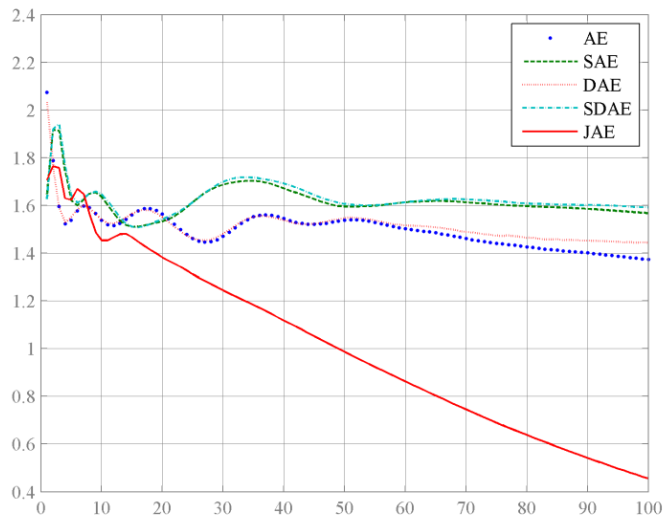

Figure 7: RMSE of $sdv$.

## 5    Conclusions

In this paper, we built a novel filter only with a single objective function, named IAE, which utilizes auxiliary information to mitigate data sparsity. The IAE achieves an appropriate balance among prediction accuracy, convergence speed, and complexity. We regard the IAE as a deep learning structure that utilizes auxiliary information for prediction. Traffic prediction, as mentioned above, is an example that demonstrates the IAE's validity. The IAE can be used to efficiently predict big data with auxiliary information in almost all fields [70, 71].

**Conflict of Interest**

*The authors declare that the research was conducted in the absence of any commercial or financial relationships that could be construed as a potential conflict of interest.*

**Author Contributions**

BX: conceptualization. BX and WP: methodology. BX and WP: software. BX and WP: validation. BX: formal analysis. WP: writing-original draft preparation. BX and WP: writing-review and editing. WP: visualization. BX: supervision, project administration, and funding acquisition.

**Funding**

The work was supported by the National Social Science Foundation of China [No. 16FJY008]; the National Natural Science Foundation of China [No. 11801060]; and the Natural Science Foundation of Shandong Province [No. ZR2016FM26].

**Acknowledgments**

The authors would like to express sincere gratitude to the referees for their valuable suggestions and comments.

**Data Availability**

The GPS trajectory dataset used to support the findings of this study are available from the corresponding author upon request, or to directly download from the following website:

https://onedrive.live.com/?authkey=%21ADgmvTgfqs4hn4Q&id=CF159105855090C5%211438&cid=CF159105855090C5.

Filter for Sparse Big Data